\pdfoutput=1 %
\documentclass[letterpaper,10pt,conference]{ieeeconf}

\IEEEoverridecommandlockouts
\makeatletter
\let\NAT@parse\undefined

\makeatother

\usepackage[numbers,sort&compress]{natbib}
\usepackage{times}
\usepackage{amsmath,amssymb,amsthm}
\usepackage{bm}
\usepackage{mathtools}
\usepackage{booktabs}
\usepackage{graphicx}
\usepackage[table]{xcolor}
\usepackage{tikz}
\definecolor{espPink}{HTML}{E89DAA}
\definecolor{promptGold}{HTML}{F1C94D}
\definecolor{promptCream}{HTML}{FBEFC7}
\colorlet{espTablePink}{espPink!20}
\usepackage{array}
\usepackage{arydshln}
\usepackage{balance}
\usepackage{xurl}
\usepackage{hyperref}
\hypersetup{hidelinks}
\usepackage{siunitx}

\newcommand{\Tref}[1]{Table~\ref{#1}}
\newcommand{\eref}[1]{Eq.~\eqref{#1}}

\newcommand{\fref}[1]{Fig.~\ref{#1}}
\newcommand{\Fref}[1]{Figure~\ref{#1}}

\newtheorem{proposition}{Proposition}
\newcommand{\R}{\mathbb{R}}
\newcommand{\E}{\mathbb{E}}
\newcommand{\norm}[1]{\left\lVert #1 \right\rVert}
\DeclareMathOperator{\ES}{ES}

\newcommand{\pnp}{\textsc{Pick \& Place}\xspace}
\newcommand{\drawer}{\textsc{Drawer Closing}\xspace}

\newsavebox{\instructionimagebox}
\newlength{\instructionpromptheight}
\newcommand{\instructionimage}[2]{%
  \sbox{\instructionimagebox}{\includegraphics[width=0.235\textwidth]{#1}}%
  \begin{tikzpicture}
    \node[inner sep=0pt, outer sep=0pt] (photo) {\phantom{\usebox{\instructionimagebox}}};
    \begin{scope}
      \clip ([yshift=\dimexpr\instructionpromptheight+0.8pt\relax]photo.south west)
        rectangle (photo.north east);
      \node[inner sep=0pt, outer sep=0pt] at (photo.center) {\usebox{\instructionimagebox}};
    \end{scope}
    \coordinate (promptwest) at ([xshift=0.2pt]photo.south west);
    \node[anchor=south west, rounded corners=1.5pt, fill=promptGold, draw=none,
      inner sep=0pt, outer sep=0pt, minimum width=10.625pt,
      minimum height=9.775pt] (prompticon) at ([yshift=1.7125pt]promptwest) {};
    \fill[promptGold]
      ([xshift=1.19pt]prompticon.south west) --
      ([xshift=1.19pt,yshift=-1.7pt]prompticon.south west) --
      ([xshift=3.485pt]prompticon.south west) -- cycle;
    \draw[white, line width=0.765pt]
      ([xshift=2.635pt,yshift=2.55pt]prompticon.south west) --
      ([xshift=6.46pt,yshift=2.55pt]prompticon.south west);
    \draw[white, line width=0.765pt]
      ([xshift=2.635pt,yshift=4.93pt]prompticon.south west) --
      ([xshift=7.285pt,yshift=4.93pt]prompticon.south west);
    \draw[white, line width=0.765pt]
      ([xshift=2.635pt,yshift=7.31pt]prompticon.south west) --
      ([xshift=8.075pt,yshift=7.31pt]prompticon.south west);
    \node[anchor=south west, rounded corners=2.1pt, draw=none, fill=promptCream,
      inner xsep=0.8pt, inner ysep=0.2pt, outer sep=0pt,
      minimum width=\dimexpr0.235\textwidth-12.5pt\relax,
      minimum height=\instructionpromptheight] at
      ([xshift=12.3pt]promptwest) {%
      \raisebox{\dimexpr(\depth-\height)/2\relax}{%
        \parbox{\dimexpr0.235\textwidth-14.3pt\relax}{%
          \centering\fontsize{5.0}{5.2}\selectfont\bfseries\texttt{#2}}%
      }%
    };
  \end{tikzpicture}%
}

\makeatletter
\def\onedot{\futurelet\@let@token\@onedot}
\def\@onedot{\ifx\@let@token.\else.\null\fi\xspace}
\makeatother
\usepackage{xspace}

\newif\ifshowcomments
\showcommentstrue %
\newcounter{todos}
\ifshowcomments
  \DeclareRobustCommand{\todo}[1]{\stepcounter{todos}\textcolor{red}{\textbf{TODO%
    \if\relax\detokenize{#1}\relax\else: #1\fi}}}
  \DeclareRobustCommand{\ym}[1]{\stepcounter{todos}\textcolor{magenta}{{[YM: #1]}}}
  \DeclareRobustCommand{\hk}[1]{\stepcounter{todos}\textcolor{blue}{{[HK: #1]}}}
  \DeclareRobustCommand{\lm}[1]{\stepcounter{todos}\textcolor{teal}{{[LM: #1]}}}
  \AtEndDocument{\ifnum\value{todos}>0
    \PackageWarning{TODOS}{There are \arabic{todos} comments left in this paper!}
  \fi}
\else
  \DeclareRobustCommand{\todo}[1]{}
  \DeclareRobustCommand{\ym}[1]{}
  \DeclareRobustCommand{\hk}[1]{}
  \DeclareRobustCommand{\lm}[1]{}
\fi

\title{\LARGE\bf ESP: Energy-Score Policy for One-Step Multimodal Action Generation}
\newif\ifanonymous
\anonymousfalse %
\ifanonymous
  \author{Anonymous Authors}
\else
\author{
\parbox{\textwidth}{\centering
Lilika Makabe$^{1,2}$\thanks{This work was done while Lilika Makabe was a research intern at Microsoft Research Asia - Tokyo.},
Heecheol Kim$^{1}$,
and Yasuyuki Matsushita$^{1}$\\[1mm]
$^{1}$Microsoft Research Asia - Tokyo\\
$^{2}$Graduate School of Information Science and Technology, the University of Osaka\\
{\tt\footnotesize
lilika.makabe@gmail.com,
\{heecheolkim, yamatsushita\}@microsoft.com}
}%
}
\fi

\newif\ifarxiv
\arxivtrue

\begin{document}
\maketitle
\ifarxiv
  \begingroup
  \renewcommand{\thefootnote}{}%
  \begin{NoHyper}%
  \footnotetext{This work has been submitted to the IEEE for possible publication. Copyright may be transferred without notice, after which this version may no longer be accessible.}%
  \end{NoHyper}%
  \endgroup
\fi

\begin{abstract}
Generative action models based on diffusion and flow matching have been increasingly adopted in vision-language-action (VLA) policies for their ability to capture diverse behaviors, including multiple valid action sequences under the same observation and instruction. Their iterative sampling procedures, however, require repeated network evaluations to generate each action chunk, increasing inference latency in closed-loop control. We propose ESP (Energy-Score Policy), a teacher-free approach that maps policy context and noise directly to an action chunk in a single network evaluation. ESP trains the action head with the energy score rather than mean squared error. Whereas squared-error regression targets the conditional mean, the energy score is strictly proper: its expected value is uniquely minimized by the target distribution. This provides a principled objective for learning multimodal action distributions without iterative sampling, with exact recovery at the population optimum when the model can represent the target distribution. Experiments on both simulation and real-world manipulation tasks demonstrate competitive task success with substantially lower action-generation latency than the flow matching baseline. These results support direct distributional learning as an efficient alternative to iterative generative robot policies.
\href{https://lilika-makabe.github.io/energy-score-policy-site/}{\textcolor{magenta}{Project page}
}

\end{abstract}

\section{Introduction}
\label{sec:intro}
Diffusion and flow matching models are increasingly used in robot policies, including vision-language-action (VLA) models, to represent diverse behaviors conditioned on observations and instructions~\citep{chi2023diffusionpolicy,black2024pi0,intelligence2025pi05}. These models can capture multiple valid action sequences for the same context rather than restrict the policy to a single deterministic prediction. Their standard sampling procedures, however, require repeated network evaluations to transform noise into an action chunk. Therefore, their action generation is inherently time-consuming, motivating faster one-step action generation with a single network evaluation.

\begin{figure}[t]
\centering
\includegraphics[width=\linewidth]{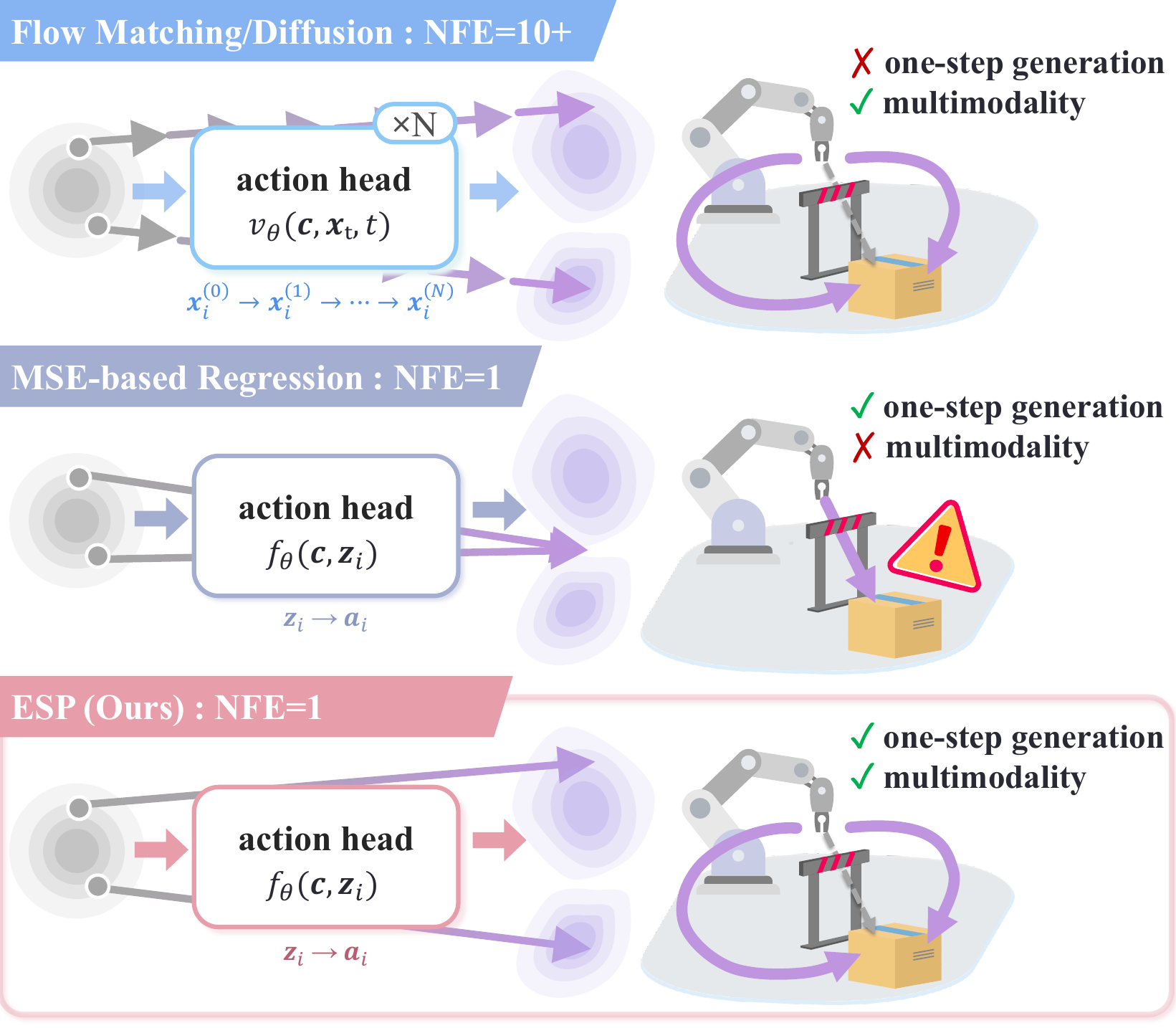}
\caption{\textbf{Energy-Score Policy (ESP) for one-step multimodal action generation.}
Flow matching can generate multimodal actions but requires integration of a velocity field over multiple network
evaluations (NFE${>}1$). Mean squared error (MSE) regression is single-step (NFE${=}1$) but collapses to the conditional mean and cannot represent multimodality. Trained with the strictly proper energy score, ESP generates
multimodal actions in one network evaluation (NFE${=}1$).
}
\vspace{-3mm}
\label{fig:teaser}
\end{figure}

We propose ESP (Energy-Score Policy), where we cast conditional action generation as strictly proper distributional regression and generate an action chunk in a single network evaluation (\fref{fig:teaser}).
The proposed ESP predicts the entire multimodal distribution of actions that are valid for a given context. We use the energy-score objective~\citep{gneiting2007strictly,bouchacourt2016disco}; in expectation over the data, it becomes smallest exactly when the generated actions match the distribution of the training data (demonstrations), and larger for every other distribution. Unlike previous one-step and few-step approaches, such as distillation and consistency-based methods~\citep{yin2024onestep,song2023consistency,zhang2025flowpolicy,chen2026meanflowvla}, ESP directly trains on the training data alone in a single stage, with no pretrained teacher, no trajectory or velocity targets, and no auxiliary critic. Different from recent teacher-free generators~\citep{deng2026drifting,gladstone2026explorative}, whose objectives can also be minimized by distributions other than the training one, ESP retains this uniqueness while keeping single-evaluation inference.

To apply ESP to an existing VLA model, we retain its pretrained vision-language backbone and policy conditioning unchanged and replace its action head with our ESP for one-step action generation. The VLM backbone can remain frozen while the ESP action-head parameters are trained without teacher distillation. For each training context, we generate several action chunks from different noise inputs, and the energy score evaluates the generated chunks against the demonstrated chunk and against one another. At inference, one single noise draw produces one action chunk in one network evaluation, without multi-step numerical integration or candidate ranking.

Experiments on synthetic regression and the LIBERO manipulation benchmark as well as real-robot experiments demonstrate that ESP can generate multimodal actions in a single network evaluation and achieves task success comparable to the flow matching baseline with substantially lower action-generation latency. These results establish energy-score training as an effective approach to fast, one-step generative action policies.

\noindent\textbf{Contributions.} Our chief contributions are:
\begin{itemize}
\item We cast conditional action generation as strictly proper distributional regression and propose an energy-score policy (ESP) for action generation.
\item Unlike diffusion/flow matching policy that requires multi-step integration, our ESP can learn the conditional noise-to-action generation in a single network evaluation, leading to faster action inference, while retaining the competitive task success rate.
\item With the same action-head architecture, in one-step, ESP achieves a competitive task success rate on LIBERO compared with the multi-step flow matching policy. Increasing the number of action-head parameters further improves the mean success rate given the same inference time budget to the flow matching policy.
\end{itemize}

\section{Related Work}
\label{sec:related}
Learning robot actions from demonstrations is a conditional distribution modeling problem: the same observation and instruction can correspond to several valid behaviors. We review multimodal prediction and generative policies, followed by objectives for efficient one-step generation.

\noindent\textbf{Multimodal prediction and generative action policies.}
Early approaches modeled multimodal conditional distributions with mixture density networks~\citep{bishop1994mixture}, conditional latent-variable models~\citep{sohn2015learning}, and conditional adversarial generators~\citep{mirza2014conditional}. Generative moment matching networks~\citep{li2015generative} train feedforward generators using Gaussian-kernel MMD (Maximum Mean Discrepancy)~\citep{dziugaite2015training}. These Gaussian-kernel MMD objectives require selecting a kernel bandwidth, whereas our energy-score objective does not require the kernel-bandwidth hyperparameter. Diffusion and flow matching subsequently emerged as expressive high-dimensional samplers that transport noise to data through a learned denoising process or vector field~\citep{ho2020denoising,song2021scorebased,lipman2023flow}.

Representing such multimodal, high-dimensional distributions is particularly relevant to robot action prediction, where the same visual observation and language instruction can correspond to several successful behaviors. Diffusion Policy~\citep{chi2023diffusionpolicy} models action sequences with conditional denoising, while more recent VLAs such as $\pi_0$~\citep{black2024pi0}, $\pi_{0.5}$~\citep{intelligence2025pi05}, GR00T N1~\citep{nvidia2025gr00t}, and FLOWER~\citep{reuss2025flower} use flow matching to generate action chunks. These methods can represent multimodal action distributions, but require multiple network evaluations. Our work preserves the stochastic \& multimodal action representation while achieving action generation only with a single network evaluation.

\noindent\textbf{One-step generation and distributional objectives.}
Reducing the cost of iterative generation has motivated distillation, consistency training, and objectives designed for few-step or one-step sampling. Distillation methods compress a multi-step model into a shorter sampler~\citep{yin2024onestep}, while consistency models can learn outputs that agree along a diffusion trajectory~\citep{song2023consistency}. In robot learning, FlowPolicy imposes self-consistency on a flow field~\citep{zhang2025flowpolicy}, MeanFlow-VLA learns an average velocity~\citep{chen2026meanflowvla}, and Let It Be Simple retains flow matching while emphasizing the high-noise regime used by one-step inference~\citep{chen2026letitbesimple}.
Drifting methods construct a training target by moving the generator's own samples toward the data with a data-adaptive drift, then train the generator to match this detached (stop-gradient) target~\citep{deng2026drifting,zuo2026driftvla}.
However, without additional assumptions, even globally minimizing the ideal population drifting loss does not guarantee that the generated distribution matches the data distribution~\citep{deng2026drifting}.

Efficiency alone, however, leaves open how a direct one-step generator should learn a multimodal distribution.
Multiple-choice learning, multiple-hypothesis prediction, and IMLE emphasize the best-matching candidates through hard or relaxed assignments~\citep{guzmanrivera2012mcl,lee2016smcl,rupprecht2017mhp,li2018implicit}. Forward XM follows this candidate-selection principle: for each training example, it evaluates the loss under several sampled noise inputs and updates the generator using the lowest-loss candidate~\citep{gladstone2026explorative}. While effective, with a fixed finite candidate budget, the hard-min endpoint-distance objective is not, in general, strictly proper: its minimum does not necessarily correspond to the data distribution.

Energy-score training instead follows the statistical lineage of proper scoring rules. The multivariate energy score combines distances to the observation with pairwise distances between generated samples. Under finite first moments, its expected value is uniquely minimized by the data distribution~\citep{gneiting2007strictly}. DISCO Nets train conditional generators using an energy-score objective and apply them to 3D hand pose estimation from depth images~\citep{bouchacourt2016disco}. We also use this established objective for one-step action generation, scoring the conditional action samples produced by the deployed action head. Minimizing its expected value recovers the target conditional distribution when the model can represent it.

\section{Method}
\label{sec:method}

\begin{figure}[t]
\centering
\includegraphics[width=\linewidth]{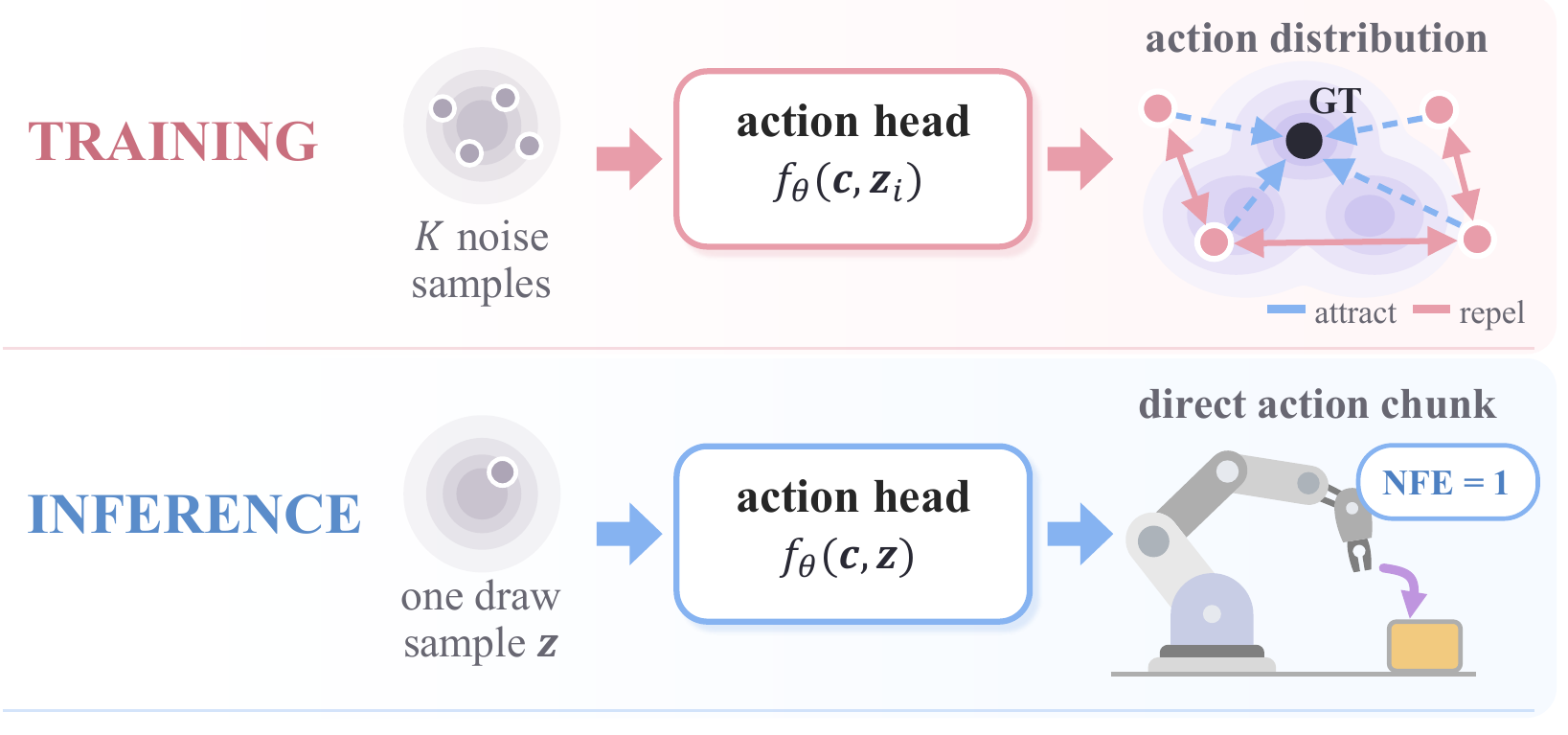}
\caption{\textbf{ESP training and one-step action generation.}
The generator maps a condition $\bm c$ and independent prior samples
$\bm z_i\sim p_z$ directly to action chunks
$\bm x_i=f_\theta(\bm c,\bm z_i)$. Training evaluates all $K$ outputs with the
energy score, whereas inference draws one $\bm z$ and requires one network evaluation.
}
\vspace{-3mm}
\label{fig:method}
\end{figure}

At each control step, we seek to sample from the demonstrated conditional action distribution
without integrating a learned sampling trajectory. Rather than learning a time-dependent
velocity field and selecting a numerical solver, we parameterize the action generator itself. We
first formalize this direct generator and explain why adding noise to a point-wise regressor is
insufficient. We then describe its energy-score training objective and one-step
sampler (\fref{fig:method}).

\subsection{Problem Setup and Pointwise Regression}
\label{sec:whyunimodal}

Let $\mathcal C$ denote the set of conditions, and let
$\bm z\sim p_z=\mathcal N(\mathbf 0,\mathbf I_{d_z})$ be an independent
noise vector in $\R^{d_z}$. The policy condition $\bm c\in\mathcal C$
is formed from encoded visual observations, a language instruction,
and robot state, while an action chunk $\bm y\in\R^d$ follows an unknown
conditional data distribution $p(\cdot\mid\bm c)$. We parameterize our ESP $f_\theta$, the one-step action generator, as
\begin{equation}
f_\theta:\mathcal C\times\R^{d_z}\to\R^d,\qquad
(\bm c,\bm z)\mapsto\bm x=f_\theta(\bm c,\bm z),
\label{eq:generator_map}
\end{equation}
where $\bm x$ is a generated action chunk in $\mathbb{R}^d$. For each fixed context $\bm c$, sampling $\bm z\sim p_z$ and computing $\bm x=f_\theta(\bm c,\bm z)$ induces a conditional predictive distribution $q_\theta$ as the push-forward of the prior $p_z$ through the generator $f_\theta$ as
\[
q_\theta(\cdot\mid\bm c)=f_\theta(\bm c,\cdot)_{\#}p_z.
\]
The learning goal is therefore optimizing the network parameters $\theta$ whose induced push-forward $q_\theta(\cdot \mid \bm c)$ reproduces the conditional data distribution $p(\cdot \mid \bm c)$,
\[
q_\theta(\cdot\mid\bm c)=p(\cdot\mid\bm c)
\]
for almost every condition~$\bm c$.

The presence of a noise input alone does not ensure that the learned
predictor achieves the distributional consistency. In fact, under a pointwise squared-error objective, variation with $\bm z$ only increases risk:
\begin{proposition}
\label{prop:mean}
For a fixed $\bm c$ with
$\E[\|\bm y\|^2\mid\bm c]<\infty$, every measurable minimizer $f:\mathcal C\times\R^{d_z}\to\R^d$ of
\[
\E_{\substack{\bm y\sim p(\cdot\mid\bm c),
              \bm z\sim p_z}}
\|f(\bm c,\bm z)-\bm y\|^2
\]
satisfies
$f(\bm c,\bm z)=\E[\bm y\mid\bm c] \overset{\mathrm{def}}{=} {\bm\mu_{\bm c}}$ almost surely, where $\bm \mu_{\bm c}$ is the conditional mean.
\end{proposition}
For a fixed context $\bm c$, the conditional MSE objective decomposes as
\begin{multline}
\E_{\substack{\bm y\sim p(\cdot\mid\bm c),
              \bm z\sim p_z}}
\|f(\bm c,\bm z)-\bm y\|^2\\
=
\E_{\bm z\sim p_z}
\|f(\bm c,\bm z)-\bm\mu_{\bm c}\|^2
+
\E_{\bm y\sim p(\cdot\mid\bm c)}
\|\bm y-\bm\mu_{\bm c}\|^2.
\label{eq:mse_decomposition}
\end{multline}
The second term on the right is independent of $f$, while the first is minimized only by mapping almost every noise sample to the conditional mean. The resulting predictive distribution is therefore the Dirac point mass
$\delta_{\bm\mu_{\bm c}}$. Thus, the target distribution can be recovered only in the deterministic
case, where each context $\bm c$ determines a unique action and $p(\cdot\mid\bm c)$ is a point mass. Otherwise, pointwise MSE cannot recover the conditional distribution.
Because the target conditional distribution is generally not a point mass but spreads over multiple valid actions, learning it requires an objective that scores the predictive distribution as a whole rather than each output independently.

\begin{figure*}[tbp]
\centering
\includegraphics[width=\textwidth]{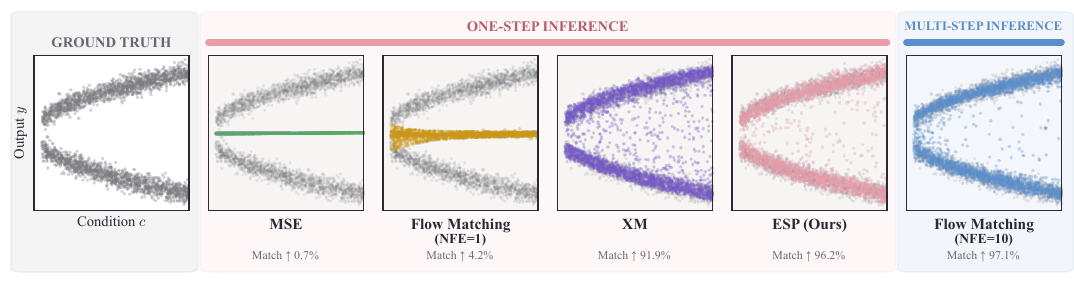}
\\ \vspace{-3mm}
\setlength{\abovecaptionskip}{0pt}
\caption{\textbf{TwoBranch simulation results.} The target is
$y=s\sqrt c+0.08\epsilon$ with equiprobable signs. The one-step group
uses one network evaluation; the multi-step reference evaluates the same trained flow model with $10$ Euler steps. Match$\uparrow$ is the percentage of predictions within $3\times$ noise standard deviations of either branch center at the same input $c$. Results use one seed and illustrate behavior rather than establishing a discriminating benchmark.
}
\label{fig:d1}
\end{figure*}

\subsection{Energy-Score Endpoint Objective}

The generator $f_\theta$ induces an implicit conditional distribution $q_\theta(\cdot \mid \bm c)$, which we can sample, but its density cannot be evaluated. We therefore train it with the energy score, which needs only samples but not likelihood. For a predictive distribution $Q$ on $\R^d$ with finite $\beta$-th moment, the energy score~\citep{gneiting2007strictly} is
\begin{equation}
\ES_\beta(Q,\bm y)
=
\E_{\bm x \sim Q}\norm{\bm x-\bm y}^{\beta}
-\tfrac12
\E_{\bm x, \bm x' \sim Q}
\norm{\bm x-\bm x'}^{\beta},
\label{eq:es}
\end{equation}
where $0<\beta<2$, and $\bm x,\bm x'$ are independent draws from $Q$.
The first term on the right measures agreement with the observed
action $\bm y$, while the second evaluates separation between two predictive
samples, $\bm x$ and $\bm x'$. Because the score depends only on samples, it applies to implicit, continuous, and discrete predictive distributions. The energy score is strictly proper: when the observation is drawn from any distribution with finite $\beta$-th moment, the expected score is uniquely minimized by predicting that same distribution. We apply this at each context, with $Q=q_\theta(\cdot \mid \bm c)$ and observations $\bm y \sim p(\cdot \mid \bm c)$.

We train the generator, ESP $f_\theta$, by minimizing the expected energy score of its induced conditional predictive distribution $q_\theta$, taken over the data distribution $\mathcal{D}$ of context-action pairs whose conditional is $p(\cdot \mid \bm c)$:
\begin{equation}
\mathcal L(\theta)
=
\E_{(\bm c, \bm y) \sim \mathcal{D}}
\left[
\ES_\beta\bigl(q_\theta(\cdot\mid\bm c),\bm y\bigr)
\right].
\label{eq:obj}
\end{equation}
Strict propriety implies that, over all predictive distributions,
this objective $\mathcal L(\theta)$ is uniquely minimized when $f_\theta$ induces $q_\theta(\cdot\mid\bm c)=p(\cdot\mid\bm c)$ for almost every $\bm c$.
This objective does not require multiple demonstrated actions for the
same condition. Each training pair $(\bm c,\bm y)$ provides one observation,
while multiple predictive samples are generated by the model.

\subsection{Training Loss}
\label{sec:fair}
The expectations in \eref{eq:es} are approximated using multiple outputs from the same
action head. For each training pair \((\bm c,\bm y)\), we draw \(K\ge2\) independent
prior samples, set \(\bm x_i=f_\theta(\bm c,\bm z_i)\), and use the standard
finite-ensemble form~\citep{ferro2014fair}
\begin{multline}
\widehat{\ES}_\beta(\{\bm x_i\},\bm y)=\frac1K\sum_{i=1}^{K}
\norm{\bm x_i-\bm y}^{\beta}\\
-\frac{1}{2K(K-1)}\sum_{i\ne j}\norm{\bm x_i-\bm x_j}^{\beta}.
\label{eq:energy_loss}
\end{multline}
Every generated action $\{\bm x_i\}$ is compared with the demonstrated action $\bm y$ and participates in the
pairwise term. We use \(\beta{=}1\) and \(K{=}8\) throughout our experiments, and optimize
\eref{eq:energy_loss} through ESP \(f_\theta\) by standard backpropagation. Regardless of the training
sample count \(K\), only one sample is drawn at inference.

\subsection{Direct One-Step Generation}
\label{sec:pushforward}
Flow and diffusion models realize a push-forward by numerically integrating a learned field. For example, rectified flow~\citep{lipman2023flow,liu2023flow,albergo2023building} introduces a time variable $t\in[0,1]$ and an intermediate state $\bm x_t \in \mathbb{R}^d$ that transports the prior sample $\bm x_0=\bm z$ into an action chunk $\bm x_1$. It learns a velocity field \(v_\theta(\bm c,\bm x, t)\) and obtains \(\bm x_1=\bm z+\int_0^1v_\theta(\bm c,\bm x_t, t)\,dt\), typically through several discrete solver steps, each requiring one network evaluation. Our model instead represents the endpoint transport directly as \(\bm x=f_\theta(\bm c,\bm z)\). At inference, one draw \(\bm z\sim p_z\) and one evaluation of \(f_\theta\) produce an action chunk, so the number of function evaluations (NFE) is one, regardless of the training sample count \(K\).

ESP's computational path does not restrict the policy to a single action per context, because multimodality is achieved by the energy-score objective used to train $f_\theta$, not by whether an action is produced in one network evaluation or by many integration steps. In principle, any conditional network that can take as input a context $\bm c$ and a noise vector $\bm z$ can serve as $f_\theta$, whether a multilayer perceptron (MLP), a convolutional network, or a transformer. Our approach can therefore be naturally applied to existing policy architectures, including VLA-based policies, without modifying their architectures. By replacing the pointwise loss with the energy-score loss, ESP enables one-step action generation while achieving performance comparable to iterative generative policies.

\section{Experiments}
\label{sec:exp}

\subsection{Evaluation Protocol}
We evaluate one-step multimodal generation through synthetic regression, VLA-based action generation on LIBERO, and real-world robot manipulation.

\noindent\textbf{Hyperparameters and comparisons.}
We compare ESP against two alternatives under matched network and training data: XM, a hard minimum-of-$K$ candidate-selection method, and native flow matching, an iterative sampler.
ESP uses \eref{eq:energy_loss} with $\beta{=}1$. Both ESP and XM generate $K{=}8$ candidates from independent noise samples during training. XM uses the objective $\mathcal{L}_\theta = \min_i\norm{f_\theta(\bm c,\bm z_i)-\bm y}$, which is a fixed-endpoint adaptation of Forward XM~\citep{gladstone2026explorative}: only the lowest-loss candidate contributes to the gradient. At inference, both methods draw one sample without candidate selection and use NFE${=}1$. Iterative flow or diffusion baselines retain their native samplers, with NFE stated for every experiment. We compare objectives through distributional metrics when the target distribution is known and through rollout success otherwise.

\subsection{Controlled Multimodal Regression}
\label{sec:lowdim}
\noindent\textbf{Dataset and implementation.}
We construct a controlled bimodal regression problem, which we call \emph{TwoBranch}, to evaluate how well each method recovers a known conditional distribution. We draw $c\sim\mathcal U[0.05,1]$ and set $y=s\sqrt c+0.08\epsilon$ with equiprobable sign $s\in\{-1,+1\}$ and $\epsilon\sim\mathcal N(0,1)$. For each input condition $c$, it produces two symmetric target branches centered at $\pm\sqrt c$.
\Fref{fig:d1} shows samples from this ground-truth distribution alongside the outputs of the trained models.
All methods use the same MLP architecture with three hidden layers of width $128$ and SiLU activations, with inputs adapted to each objective.
We train them for $30$k iterations at a constant learning rate of $2{\times}10^{-4}$.
We evaluate the same trained flow matching model with either one or ten Euler steps.
We count a prediction as a match if its output differs from either center by at most
$0.24$, which is $3\times$ the data noise standard deviation of $0.08$.
\emph{Match} is the percentage of predictions that satisfy this criterion, and we use it as the distributional-fidelity metric.

\noindent\textbf{Results.}
MSE regression concentrates its predictions at the conditional mean rather than
the two target branches, consistent with the characterization in
Proposition~\ref{prop:mean}.
Flow matching learns a velocity field that is integrated to generate samples.
With only one step, its predictions also concentrate between the two branches.
MSE and one-step flow obtain Match rates of $0.7\%$ and $4.2\%$, respectively.
ESP and XM both generate the two branches in one network evaluation, but XM produces
more outliers in the low-density region between them. ESP reaches a Match rate of $96.2\%$, comparable to $91.9\%$ of XM and $97.1\%$ of ten Euler step flow matching.
Thus, directly training the endpoint generator with the energy score captures both
branches and approaches the Match rate of iterative flow without repeated network evaluations.

\begin{figure*}[t]
\centering
\raisebox{-0.538767\height}{%
\includegraphics[width=0.48\textwidth]{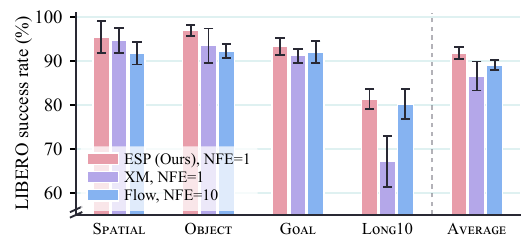}%
}\hfill
\raisebox{\dimexpr(\depth-\height)/2\relax}{%
\begin{minipage}[c]{0.49\textwidth}
\centering
\footnotesize
\begingroup
\setlength{\aboverulesep}{0pt}
\setlength{\belowrulesep}{0pt}
\setlength{\tabcolsep}{3.5pt}
\renewcommand{\arraystretch}{1.35}
\begin{tabular*}{\linewidth}{@{\extracolsep{\fill}}l>{\columncolor{espTablePink}}ccc@{}}
\toprule
Suite & \shortstack{\rule{0pt}{3.6ex}ESP (Ours)\\NFE${=}1$} & \shortstack{\rule{0pt}{3.6ex}XM\\NFE${=}1$} & \shortstack{\rule{0pt}{3.6ex}Flow\\NFE${=}10$} \\
\midrule
\textsc{Spatial} & $\mathbf{0.955{\pm}0.036}$ & $0.947{\pm}0.028$ & $0.918{\pm}0.026$ \\
\textsc{Object} & $\mathbf{0.970{\pm}0.013}$ & $0.935{\pm}0.039$ & $0.923{\pm}0.016$ \\
\textsc{Goal} & $\mathbf{0.933{\pm}0.019}$ & $0.912{\pm}0.016$ & $0.920{\pm}0.025$ \\
\textsc{Long10} & $\mathbf{0.813{\pm}0.023}$ & $0.672{\pm}0.058$ & $0.802{\pm}0.034$ \\
\midrule
\textbf{Average} & $\mathbf{0.918{\pm}0.014}$ & $0.866{\pm}0.033$ & $0.891{\pm}0.012$ \\
\bottomrule
\end{tabular*}
\endgroup
\end{minipage}%
}
\caption{Suite-wise $\pi_{0.5}$ closed-loop success on LIBERO.
Left: mean success rates in percent, with sample SD error bars.
Right: the same results as success fractions (mean $\pm$ SD).
Statistics use three training seeds; Average first averages the four suites within each seed.}
\label{fig:pi05_headline}
\end{figure*}

\subsection{\texorpdfstring{$\pi_{0.5}$}{pi0.5} on LIBERO}
\label{sec:pi05}
$\pi_{0.5}$~\citep{intelligence2025pi05} is a VLA policy with a flow matching action expert that requires multiple Euler steps at inference. We evaluate whether replacing this iterative decoder with a one-step ESP preserves closed-loop task success.

\noindent\textbf{Model adaptation.} We freeze the $\pi_{0.5}$ vision-language backbone
and retain its tokenization and conditioning. We train only the action-side modules
and replace the flow matching action head with a one-step ESP or XM head
$f_\theta(\bm c,\bm z)$. Since the architecture assumes diffusion time-step $t$ as an additional input, we simply fix it to $t{=}1$ during the training and inference time for both ESP and XM.
All methods train from scratch on the action side with random initialization.
The optimizer, schedule, and initialization are shared, so the substitution isolates the action-head
objective and its sampler.

\noindent\textbf{Training and evaluation.}
We train on the \textsc{Spatial}, \textsc{Object}, \textsc{Goal},
and \textsc{Long10} suites of LIBERO~\citep{liu2024libero}
with batch size $32$, AdamW, global gradient clipping at $1.0$, and EMA decay $0.999$.
We use a linear warm-up from approximately $5{\times}10^{-9}$ to $5{\times}10^{-5}$ over the first 10,000 optimization steps, after which the learning rate remains constant.

At each generation, the policy predicts ten actions as a chunk and executes the first five.
All models have the same number of trainable parameters and use the same base
checkpoint for the frozen backbone.
We train each model for $60$k steps with seeds 42, 43, and 44, and evaluate the last checkpoint using the fixed evaluation protocol. Each suite contains ten tasks with 20 rollouts per task and seed, totaling 2,400 rollouts per head.

\noindent\textbf{Closed-loop success.}
We show the suite-wise LIBERO success rates at $60$k training iterations in \fref{fig:pi05_headline}. ESP exhibits competitive success rates over the ten-step flow sampler. Its advantage over XM under the same one-step architecture further supports energy scoring over hard-min candidate assignment for learning effective action policies.

\noindent\textbf{Latency.} \Tref{tab:pi05_latency} compares inference latency across methods. We measure the latency using $\pi_{0.5}$ on a single A100 GPU with batch size one and action horizon $H{=}10$, using $15$ warm-up and $100$ timed calls per path. Cached inference reuses the KV cache in the VLM; uncached inference includes its computation.

\begin{table}[tbp]
\centering
\small
\caption{Policy inference latency (ms)}
\vspace{-2mm}
\label{tab:pi05_latency}
\begingroup
\setlength{\aboverulesep}{0pt}
\setlength{\belowrulesep}{0pt}
\renewcommand{\arraystretch}{1.25}
\resizebox{\columnwidth}{!}{%
\begin{tabular}{l>{\columncolor{espTablePink}}ccc}
\toprule
Path & ESP,~NFE${=}1$ & XM,~NFE${=}1$ & Flow,~NFE${=}10$ \\
\midrule
Action expert             & $3.35$ & $3.31$ & $28.39$ \\
Cached end to end         & $4.46$ & $4.22$ & $29.14$ \\
Uncached end to end       & $59.01$ & $58.78$ & $82.54$ \\
\bottomrule
\end{tabular}}
\endgroup
\vspace{-3mm}
\end{table}

\begin{figure*}[tbp]
\centering
\includegraphics[width=\linewidth,height=0.82\textheight,keepaspectratio]{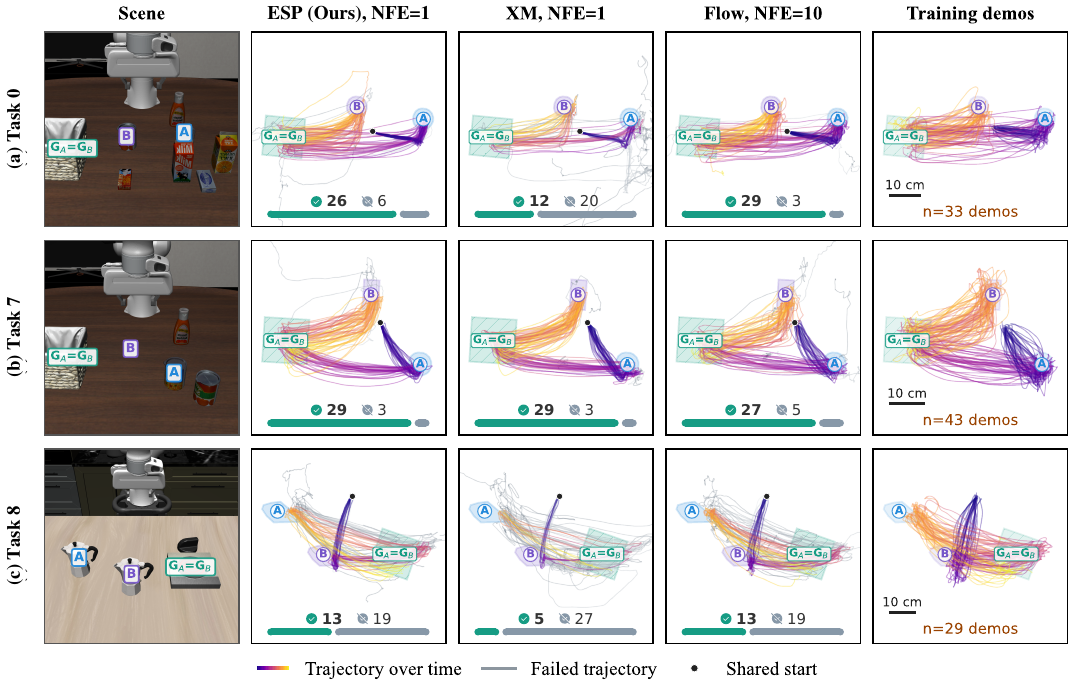}\\ \vspace{-4mm}
\caption{\textbf{Fixed-initialization end-effector $XY$ trajectories.}
Rows show Tasks 0, 7, and 8 from the six-task experiment.
Columns show the scene, ESP and XM at NFE${=}1$, flow at NFE${=}10$, and demonstrations.
Scene labels show objects A, B and placement goals $G_A, G_B$, and obstacles visible in the scene are unlabeled. Each policy subfigure shows all 32 rollouts. The success/failure bar shows the success rate across rollouts with different policy-noise sequences. Successful trajectories progress from purple to yellow over time, and failures are gray. The black dot marks the shared rollout start.
Task 8 success additionally requires a non-spatial goal condition (the stove-on predicate), which is not visible in the $XY$ paths.
}
\label{fig:action_dispersion}
\end{figure*}

ESP reduces action-expert latency from $28.39$ ms to $3.35$ ms, an $8.47\times$ speedup over native flow matching. Cached end-to-end inference is $6.53\times$ faster. Even with VLM computation included, ESP reduces end-to-end latency from $82.54$ ms to $59.01$ ms, retaining a $1.40\times$ speedup. Thus, one-step action generation accelerates both the action expert and the complete policy.

\subsection{Action-Expert Capacity}
\label{sec:ffn_capacity}
One-step generation reduces the computation spent on repeated network evaluations during action generation. We examine whether this saving can instead support a larger action expert. The expanded ESP increases the action-expert feed-forward network (FFN) width from
$4096$ to $73728$ while retaining the frozen vision-language backbone, action-expert
hidden size, attention heads, KV cache structure, action horizon $H{=}10$, and output shape.
The wider model has 7.204B parameters, including 4.280B on the action side and an unchanged
2.923B vision-language backbone, with totals rounded independently.

\Tref{tab:ffn_capacity} reports LIBERO success for training seeds 42, 43, and 44, with 200 episodes per suite and seed per model. Both models use NFE${=}1$. The baseline ESP~(Ours) is our results with the same architecture as native flow matching, and ESP~(Ours) expanded is ours with the larger FFN width.

\begin{table}[tbp]
\centering
\small
\vspace{2mm}
\caption{Action-expert capacity ablation within ESP\\(success rate mean $\pm$ SD)}
\vspace{-2mm}
\label{tab:ffn_capacity}
\begin{tabular*}{\columnwidth}{@{\extracolsep{\fill}}lcc@{}}
\toprule
 & ESP (Ours) & ESP (Ours), expanded \\
\midrule
FFN width & 4096 & 73728 \\
\midrule
\textsc{Spatial} & $\mathbf{0.955\pm0.036}$ & $0.953\pm0.019$ \\
\textsc{Object} & $0.970\pm0.013$ & $\mathbf{0.985\pm0.009}$ \\
\textsc{Goal} & $0.933\pm0.019$ & $\mathbf{0.953\pm0.008}$ \\
\textsc{Long10} & $0.813\pm0.023$ & $\mathbf{0.857\pm0.012}$ \\
Average & $0.918\pm0.014$ & $\mathbf{0.937\pm0.004}$ \\
\midrule
Latency (ms) & $3.35$ & $21.15$ \\
Time / Flow time & $0.118$ & $0.732$ \\
\bottomrule
\end{tabular*}
\vspace{-2mm}
\end{table}

The expanded model reaches $0.9371\pm0.0036$ mean success rate $\pm$ SD, compared with $0.9179\pm0.0140$ for baseline ESP. Overall, the mean success rate is higher or comparable on all four suites, while achieving lower SD, more stable results.
Our one-step generation head can either achieve the faster inference time, or enable larger action capacity given the same latency budget.

\subsection{Multimodality Ablation}
\label{sec:dispersion}
We evaluate whether each trained action head generates diverse trajectories from
the same initial state and how reliably these trajectories complete the task.

\noindent\textbf{Protocol.}
To examine how input noise affects the generated trajectories, we fix the simulator initial state for each task and perform 32 rollouts with different noise sequences. We evaluate the six LIBERO-10 tasks that require placing two movable objects at their respective goals: Tasks 0, 1, 4, 6, 7, and 8. All methods use training seed 42 and the checkpoint after $60$k training iterations, with the same noise schedule across methods. We concatenate the executed end-effector positions to obtain each rollout trajectory.

\noindent\textbf{Task success and trajectory spread.}
\Fref{fig:action_dispersion} visualizes Tasks 0, 7, and 8 from this experiment; \Tref{tab:fixed_init_diagnostic} summarizes all six tasks. ESP achieves more consistent task success, succeeding in 162 of 192 trials,
compared with 127 for XM and 155 for flow.
We quantify the spread of successful trajectories using the mean RMS symmetric
Chamfer distance over distinct pairs of end-effector XYZ curves.
We compute this metric for each task and report the averaged scores.

\begin{table}[tbp]
\centering
\small
\caption{Fixed-initialization stochastic rollout diagnostic}
\label{tab:fixed_init_diagnostic}
\begingroup
\setlength{\aboverulesep}{0pt}
\setlength{\belowrulesep}{0pt}
\renewcommand{\arraystretch}{1.25}
\begin{tabular*}{\columnwidth}{@{\extracolsep{\fill}}l>{\columncolor{espTablePink}}ccc@{}}
\toprule
Metric/Method & ESP & XM & Flow \\
 & NFE${=}1$ & NFE${=}1$ & NFE${=}10$ \\
\midrule
Success rate & $\mathbf{162/192}$ & $127/192$ & $155/192$ \\
Chamfer$\uparrow$ (cm) & $2.122$ & $1.965$ & $\mathbf{2.280}$ \\
\bottomrule
\end{tabular*}
\endgroup
\vspace{-2mm}
\end{table}

The mean spread is $2.122$ cm for ESP and $1.965$ cm for XM, a difference of $0.157$ cm.
As shown in \fref{fig:action_dispersion}, ESP generates more diverse successful
trajectories than XM, with diversity similar to that of flow matching.

\begin{figure*}[!t]
\centering
\begingroup
\setlength{\tabcolsep}{0pt}
\begin{tabular*}{\textwidth}{@{\extracolsep{\fill}}cccc@{}}
{\footnotesize\textbf{Environment}} &
{\footnotesize\textbf{\drawer}} &
{\footnotesize\textbf{\pnp}} & \\[1mm]
\raisebox{-0.5\height}{\includegraphics[width=0.235\textwidth]{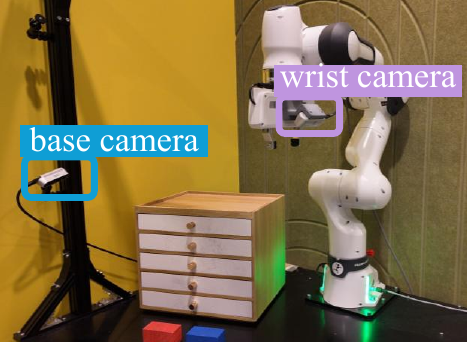}} &
\raisebox{-0.5\height}{\instructionimage{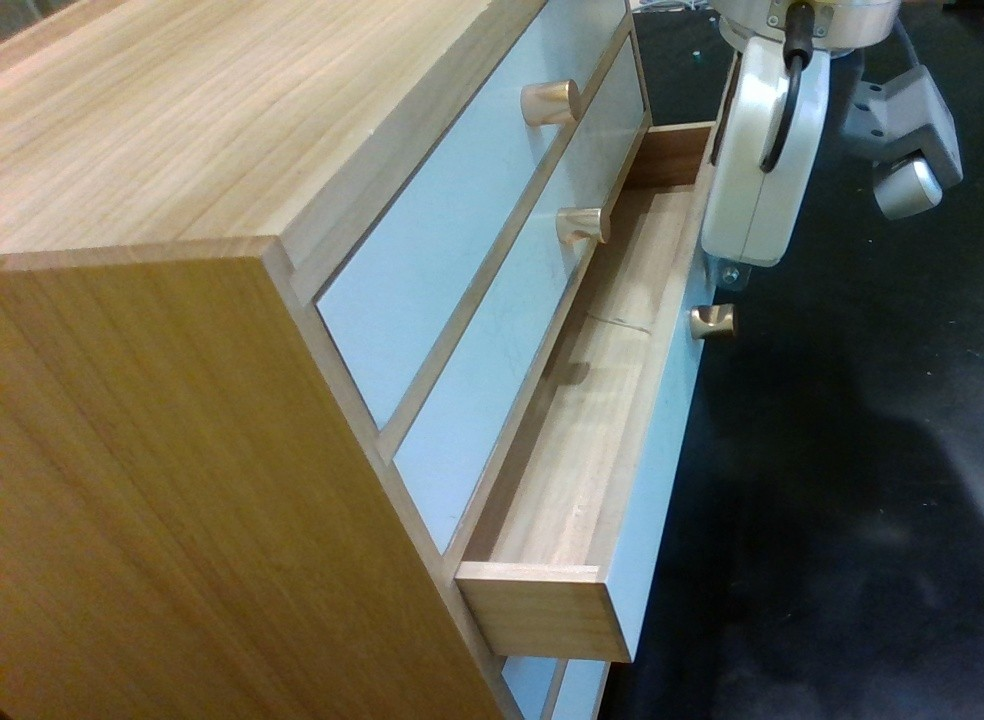}{``close the drawer''}} &
\raisebox{-0.5\height}{\instructionimage{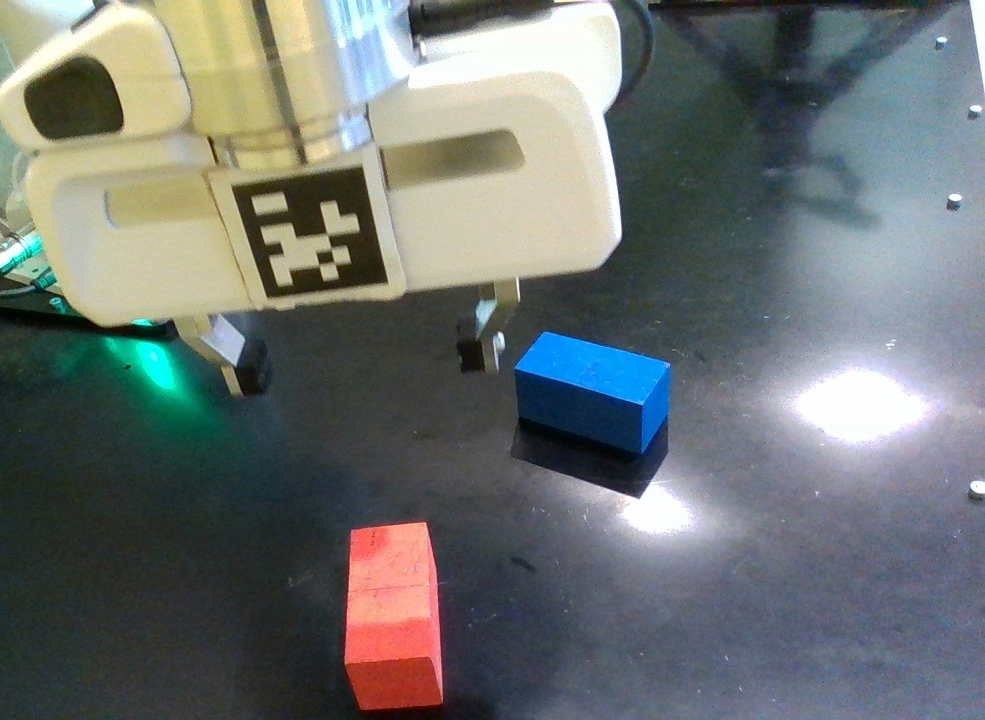}{``pick up the block and\\place it on the other block''}} &
\raisebox{\dimexpr(\depth-\height)/2\relax}{%
\begin{minipage}{0.27\textwidth}
\centering
\scriptsize
\raisebox{5pt}{{\footnotesize\textbf{Success Rate}}}\par\vspace{1pt}
\renewcommand{\arraystretch}{1.35}
\setlength{\dashlinedash}{2pt}
\setlength{\dashlinegap}{1.5pt}
\setlength{\tabcolsep}{3pt}
\setlength{\aboverulesep}{0pt}
\setlength{\belowrulesep}{0pt}
\begin{tabular}{lc>{\columncolor{espTablePink}}c}
\toprule
Task & \shortstack{\rule{0pt}{2.6ex}Flow\\NFE=4} & \shortstack{\rule{0pt}{2.6ex}ESP (Ours)\\NFE=1} \\
\midrule
\drawer & 8/15 & $\mathbf{12}/15$ \\
\hdashline
\pnp & & \\
\quad Pick-only & 9/15 & $\mathbf{10}/15$ \\
\quad Full task & 1/15 & $\mathbf{7}/15$ \\
\bottomrule
\end{tabular}
\end{minipage}%
}
\end{tabular*}
\endgroup
\par\vspace{4mm}
\includegraphics[width=\textwidth]{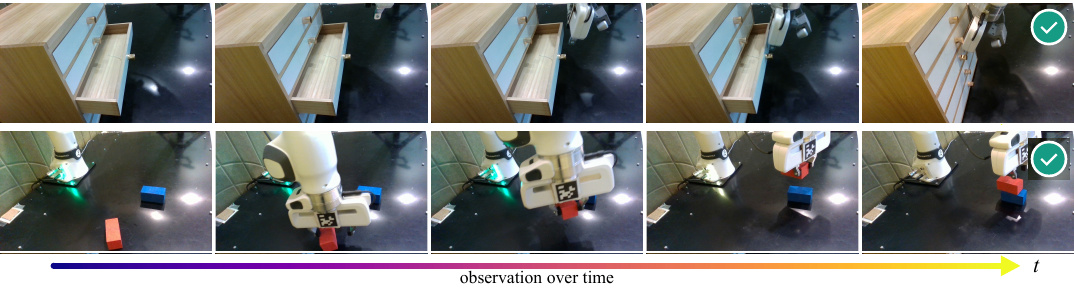}\\
\vspace{-3mm}
\caption{\textbf{Franka environment, tasks, language instructions, and physical rollout outcomes.}
The environment figure shows the base and wrist cameras, task figures show representative drawer and pick~\&~place examples from the physical policy rollouts. The bottom two strips show observations over time from successful \drawer (upper) and \pnp (lower) rollouts. The table reports successful trials / total trials.}
\label{fig:franka_domains}
\end{figure*}

\subsection{Real-World Manipulation with GR00T}
\label{sec:franka}
\noindent\textbf{Experimental setup.}
We evaluate two real-world manipulation tasks using a Franka Research 3:
\pnp and \drawer.
To verify that the proposed approach generalizes across policy architectures, we conduct the real-world experiments using GR00T as the backbone VLA.
For each task, we train on
approximately 30 to 40 human-teleoperated demonstrations.
\Fref{fig:franka_domains}~(top left) shows the experimental setup in the real world. The policy receives RGB
images from base-mounted and wrist-mounted Intel RealSense D435 cameras, together with a language instruction and
a current robot state, and predicts the future actions.

\noindent\textbf{Policy and baselines.}
Both ESP and Flow start from GR00T-N1.7-3B. We freeze the vision-language backbone
and only fine-tune the pretrained action head.
Both the predicted and executed action
chunks contain eight actions. Flow uses the native flow matching
objective and the default four-step sampler of GR00T-N1.7-3B (NFE${=}4$).
Both methods are trained for $60$k  iterations for \drawer and $100$k for \pnp with a batch size of eight and a constant
learning rate of $2.5{\times}10^{-5}$ without warm-up. We use weight decay $10^{-5}$
and training seed 42.

\noindent\textbf{Evaluation.}
We compare closed-loop task success in the real-world experiment with task-specific success criteria.\\
\noindent\drawer: We open the drawer from approximately $\SI{7}{\centi\metre}$ to $\SI{12}{\centi\metre}$ before fully closed, with perturbations applied to both the drawer position and the initial robot pose. A rollout is considered successful if the drawer is closed to within $\SI{5}{\milli\metre}$ of the fully closed position. \\
\noindent\pnp:
The two blocks are placed with random perturbations, and the robot is tasked with stacking one block on top of the other. We report the overall task success and the pick-only success in the task. A rollout is considered successful for picking if the target block is lifted from the table, and for the full task if it is stacked on the other block without toppling.
A rollout is terminated and counted as a failure if it exceeds \SI{180}{\second} or if excessive contact forces are detected, such as the end-effector colliding with the table.

\noindent\textbf{Results.}
The bottom two strips of \fref{fig:franka_domains} show sequences of observations
from successful \drawer and \pnp rollouts, providing qualitative
examples of policy execution on the physical robot. The top right table shows the success rate in all tasks. Ours achieves overall success rate gains across all three tasks, and continues to outperform the baseline in real-world experiments.

\begin{figure}[t]
\centering
\includegraphics[width=\columnwidth]{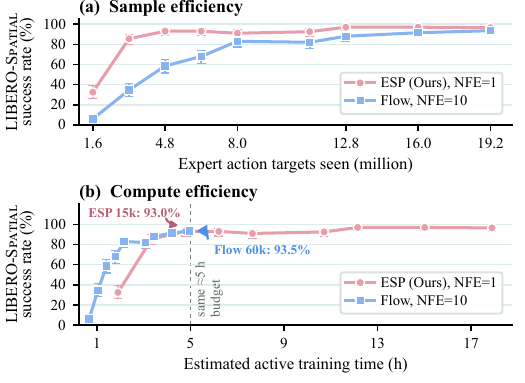}\\
\caption{
\textbf{Training progress on LIBERO-\textsc{Spatial} ($K{=}8$).}
Success rate over (a) expert-action exposure and (b) training time for checkpoints
from the same runs as the main $\pi_{0.5}$ results. Although flow matching processes
optimizer steps $3.81\times$ faster, ESP at 15k steps (4.78h) achieves $93.0\%$ success,
comparable to $93.5\%$ for flow matching at 60k steps (4.98h).}
\label{fig:training_progress}
\end{figure}

\section{Limitations}
The main limitation of ESP is its training cost. Because the energy score compares samples, each update draws $K$ noise-conditioned candidates per example instead of one, and its pairwise repulsion term evaluates $O(K^2)$ sample distances, with both the forward and backward passes running over all $K$ candidates. This increases computation and activation memory compared to single-sample flow matching training. Indeed, during $\pi_{0.5}$ training, each flow matching optimizer step runs about $3.81 \times$ faster than an ESP step when $K{=}8$.

The per-step overhead is partly offset by sample efficiency. As \fref{fig:training_progress} shows, ESP reaches a success rate comparable to flow matching ($93.0\%$ vs. $93.5\%$) using only about a quarter as many optimizer steps and expert actions, so the two reach comparable success in nearly the same wall-clock time (ESP at 15k steps in 4.78h vs. flow matching at 60k steps in 4.98h). Although this sample efficiency compensates for much of the training-time overhead in this experiment, reducing the training overhead remains an important direction for future work. ESP's computational advantage instead appears at deployment, where it generates each action chunk in a single network evaluation (NFE${=}1$).

\section{Conclusion}
We studied conditional action generation as a direct noise-to-action map rather than an integrated trajectory. %
The energy score trains the generator to match the full conditional distribution from demonstrations with a single target per context.
Building on this, we introduced ESP, a one-step generative action policy that directly maps policy context and noise to an action chunk in a single forward pass. Across simulated and real-world manipulation tasks, ESP achieves competitive task success while substantially reducing action-generation latency compared to flow matching policies that require iterative sampling. These results demonstrate that ESP can be an efficient alternative to iterative generative policies for robotic control. Beyond reducing latency, our results suggest a different allocation of computational resources for generative robot policies. Eliminating iterative sampling allows the same time budget to be used for larger action heads rather than repeated evaluations of a smaller model. The improved performance of wider one-step action heads indicates that, under practical control-frequency constraints, increasing policy capacity may be a more effective use of compute than iterative action generation.

\section*{Acknowledgment}
We thank Namiko Saito, Seongheon Park, Shulin Tian, and Zeyi Huang for helpful discussions and valuable feedback throughout the project.
We used GitHub Copilot for text editing and LaTeX assistance. All scientific
claims, text, experimental results, and conclusions were determined and verified by the authors.

\balance
{\footnotesize
\bibliographystyle{IEEEtran}
\bibliography{references}
}

\end{document}